\documentclass{article}

\usepackage[final]{neurips_2019}

\usepackage[utf8]{inputenc} % allow utf-8 input
\usepackage[T1]{fontenc}    % use 8-bit T1 fonts
\usepackage{hyperref}
\usepackage{url}            % simple URL typesetting
\usepackage{booktabs}       % professional-quality tables
\usepackage{amsfonts}       % blackboard math symbols
\usepackage{nicefrac}       % compact symbols for 1/2, etc.
\usepackage{microtype}      % microtypography
\usepackage[ruled]{algorithm2e}
\usepackage{graphicx}
\usepackage{lipsum}
\usepackage{amsfonts}
\usepackage{amsmath}
\usepackage[x11names]{xcolor}
\usepackage{wrapfig}
\usepackage{algorithmic, multicol}

\definecolor{myBlue}{rgb}{0.1,0.1,0.5}

\hypersetup{
	colorlinks=true,
	linkcolor=blue,
	filecolor=magenta,      
	urlcolor=myBlue, 	
	citecolor=myBlue, 	
}

\newcommand{\repparams}{\theta}	
\newcommand{\taskparams}{W}	
\newcommand{\repdim}{d}
\newcommand{\numupdates}{k}
	
\newcommand{\loss}{\ell}	
\newcommand{\yhat}{\hat{y}}	

\newcommand{\defeq}{\mathrel{\overset{\makebox[0pt]{\mbox{\normalfont\tiny\sffamily def}}}{=}}}

\newcommand{\CLP}{\text{CLP}}	
\newcommand{\OML}{\text{OML}}

\newcommand{\task}{\mathcal{T}}
\newcommand{\inp}{X}

\newcommand{\encoder}{\phi_{\repparams}}

\title{Meta-Learning Representations for \\ Continual Learning}

\author{%
	Khurram Javed, Martha White\\
	Department of Computing Science\\
	University of Alberta\\
	T6G 1P8 \\
	\texttt{kjaved@ualberta.ca, whitem@ualberta.ca} 
}

\begin{document}	
	\maketitle
	\begin{abstract}
		A continual learning agent should be able to build on top of existing knowledge to learn on new data quickly while minimizing forgetting. Current intelligent systems based on neural network function approximators arguably do the opposite---they are highly prone to forgetting and rarely trained to facilitate future learning. One reason for this poor behavior is that they learn from a representation that is not explicitly trained for these two goals. In this paper, we propose \OML, an objective that directly minimizes catastrophic interference by learning representations that accelerate future learning and are robust to forgetting under online updates in continual learning. We show that it is possible to learn naturally sparse representations that are more effective for online updating. Moreover, our algorithm is complementary to existing continual learning strategies, such as MER and GEM. Finally, we demonstrate that a basic online updating strategy on representations learned by \OML\ is competitive with rehearsal based methods for continual learning. \footnote{We release an implementation of our method at \url{https://github.com/khurramjaved96/mrcl}}
	\end{abstract}	

	\section{Introduction} 
	
	Continual learning---also called cumulative learning and lifelong learning---is the problem setting where an agent faces a continual stream of data, and must continually make and learn new predictions. The two main goals of continual learning are (1) to exploit existing knowledge of the world to quickly learn predictions on new samples (accelerate future learning) and (2) reduce interference in updates, particularly avoiding overwriting older knowledge. Humans, as intelligence agents, are capable of doing both. For instance, an experienced programmer can learn a new programming language significantly faster than someone who has never programmed before and does not need to forget the old language to learn the new one. Current state-of-the-art learning systems, on the other hand, struggle with both \citep{french1999catastrophic,kirkpatrick2017overcoming}.  
	
	Several methods have been proposed to address catastrophic interference. These can generally be categorized into methods that (1) modify the online update to retain knowledge, (2) replay or generate samples for more updates and (3) use semi-distributed representations. Knowledge retention methods prevent important weights from changing too much, by introducing a regularization term for each parameter weighted by its importance \citep{kirkpatrick2017overcoming,aljundi2018memory,zenke2017continual,lee2017overcoming,liu2018rotate}. Rehearsal methods interleave online updates with updates on samples from a model. Samples from a model can be obtained by replaying samples from older data \citep{lin1992self,mnih2015human, chaudhry2018efficient, riemer2018learning, rebuffi2017icarl, lopez2017gradient, aljundi19}, by using a generative model learned on previous data \citep{sutton1990integrated,shin2017continual}, or using knowledge distillation which generates targets using predictions from an older predictor \citep{li2018learning}. These ideas are all complementary to that of learning representations that are suitable for online updating.

	Early work on catastrophic interference focused on learning semi-distributed (also called sparse) representations \citep{french1991using,french1999catastrophic}. Recent work has revisited the utility of sparse representations for mitigating interference \citep{liu2018utility} and for using model capacity more conservatively to leave room for future learning \citep{aljundi2018selfless}. These methods, however, use sparsity as a proxy, which alone does not guarantee robustness to interference. A recently proposed online update for neural networks implicitly learns representations to obtain non-interfering updates \citep{riemer2018learning}. Their objective maximizes the dot product between gradients computed for different samples. The idea is to encourage the network to reach an area in the parameter space where updates to the entire network have minimal interference and positive generalization. This idea is powerful: to specify an objective to explicitly mitigate interference---rather than implicitly with sparse representations. 

% One outstanding issue is that the representation itself is difficult to learn online, and suffers from interference. The above methods learn online directly on potentially dense observations, such as raw pixels. A deep neural network updated online using gradient descent with dense inputs is bound to suffer from catastrophic forgetting. This is because initial layers of deep convolutional neural networks extract low-level features useful for many upstream tasks \citep{farabet2012learning}. In the absence of iid sampling, a gradient descent update with dense inputs greedily changes these initial layers to extract features for the current task only. Moreover, these changes to the initial layers have compounding effects on the prediction as the input distribution of later layers is also changed. Current state-of-the-art continual learning systems avoid this problem by obtaining iid samples from an experience replay buffer \citep{chaudhry2019continual,riemer2018learning, rebuffi2017icarl, lopez2017gradient, aljundi19}. Such an approach, however, is not scalable for continual learning. 

\begin{figure}
	\centering
   	\includegraphics[width=0.80	\linewidth]{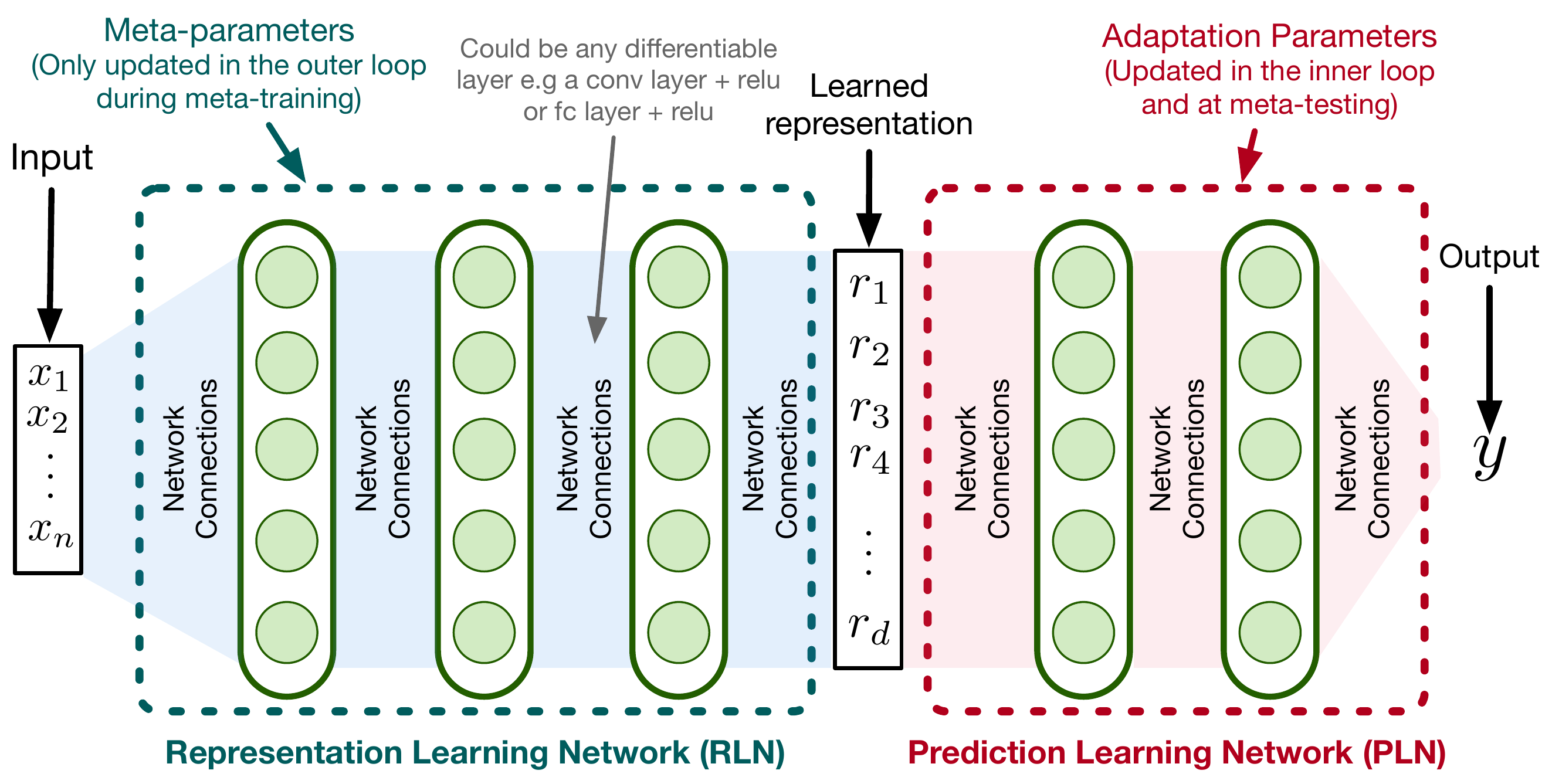}
	\caption{An example of our proposed architecture for learning representations for continual learning. 
	During the inner gradient steps for computing the meta-objective, we only update the parameters in the prediction learning network (PLN). 
	We then update both the representation learning network (RLN) and the prediction learning network (PLN) by taking a gradient step with respect to our meta-objective. The online updates for continual learning also only modify the PLN. Both RLN and PLN can be arbitrary models. 
	}
	\label{fig:overview}
\end{figure}
In this work, we propose to explicitly learn a representation for continual learning that avoids interference and promotes future learning. We propose to train the representation with $\OML$ -- a meta-objective that uses catastrophic interference as a training signal by directly optimizing through an online update. The goal is to learn a representation such that the stochastic online updates the agent will use at meta-test time improve the accuracy of its predictions in general. We show that using our objective, it is possible to learn representations that are more effective for online updating in sequential regression and classification problems. Moreover, these representations are naturally highly sparse. Finally, we show that existing continual learning strategies, like Meta Experience Replay \citep{riemer2018learning}, can learn more effectively from these representations. 

\section{Problem Formulation}
A Continual Learning Prediction (CLP) problem consists of an unending stream of samples
\begin{equation*}
 \task = (X_1, Y_1), (X_2, Y_2), \ldots, (X_t, Y_t), \ldots
\end{equation*}
for inputs $X_t$ and prediction targets $Y_t$, from sets $\mathcal{X}$ and $\mathcal{Y}$ respectively.\footnote{This definition encompasses the continual learning problem where the tuples also include task descriptors $T_t$ \citep{lopez2017gradient}. $T_t$ in the tuple $(X_t, T_t, Y_t)$ can simply be considered as part of the inputs.}   
The random vector  $Y_t$ is sampled according to an unknown distribution $p(Y | X_t)$.
We assume the process $X_1, X_2, \ldots, X_t, \ldots$ has a marginal distribution $\mu: \mathcal{X} \rightarrow [0, \infty)$, that reflects how often each input is observed. 
This assumption allows for a variety of correlated sequences. For example, $X_t$ could be sampled from a distribution potentially dependent on past variables $X_{t-1}$ and $X_{t-2}$. The targets $Y_t$, however, are dependent only on $X_t$, and not on past $X_i$. 
We define $\mathcal{S}_k = (X_{j+1} Y_{j+1}), (X_{j+2} Y_{j+2}) \ldots, (X_{j+k}, Y_{j+k})$, a random trajectory of length $k$ sampled from the $\CLP$ problem $\task$. Finally, $p(S_k | \task)$ gives a distribution over all trajectories of length $k$ that can be sampled from problem $\task$. 
% The goal of the agent is to learn a function $f: \mathcal{X} \rightarrow \mathcal{Y}$, to predict targets $y$ from inputs $x$ using $\mathcal{S}_k$ as training data. 

 For a given CLP problem, our goal is to learn a function $f_{\taskparams, \repparams}$ that can predict $Y_t$ given $X_t$. More concretely, let $\loss: \mathcal{Y} \times \mathcal{Y} \rightarrow \mathbb{R}$ be the function that defines loss between a prediction $\yhat \in \mathcal{Y}$ and target $y$ as $\loss(\yhat, y)$. If we assume that inputs $X$ are seen proportionally to some density $\mu: \mathcal{X} \rightarrow [0, \infty)$, then we want to minimize the following objective for a CLP problem: 
\begin{equation}
\mathcal L_{CLP}(\taskparams, \repparams) 
\defeq \mathbb{E}[ \loss(f_{\taskparams, \repparams}(X), Y) ] 
= \int \left[\int \loss(f_{\taskparams, \repparams}(x), y) p(y|x) dy \right] \mu(x) dx 
.
\end{equation}
where $\taskparams$ and $\repparams$ represent the set of parameters that are updated to minimize the objective. To minimize $\mathcal L_{CLP}$, we limit ourselves to learning by online updates on a single $k$ length trajectory sampled from $p(S_k | \task)$. This changes the learning problem from the standard iid setting -- the agent sees a single trajectory of correlated samples of length $k$, rather than getting to directly sample from $p(x, y) = p(y|x) \mu(x)$. This modification can cause significant issues when simply applying standard algorithms for the iid setting. Instead, we need to design algorithms that take this correlation into account.

A variety of continual problems can be represented by this formulation. One example is an online regression problem, such as predicting the next spatial location for a robot given the current location; another is the existing incremental classification benchmarks. The CLP formulation also allows for targets $Y_t$ that are dependent on a history of the most recent $m$ observations. This can be obtained by defining each $X_t$ to be the last $m$ observations. The overlap between $X_t$ and $X_{t-1}$ does not violate the assumptions on the correlated sequence of inputs. Finally, the prediction problem in reinforcement learning---predicting the value of a policy from a state---can be represented by considering the inputs $X_t$ to be states and the targets to be sampled returns or bootstrapped targets.  

\begin{figure}
	\centering
	\includegraphics[width=0.60\linewidth]{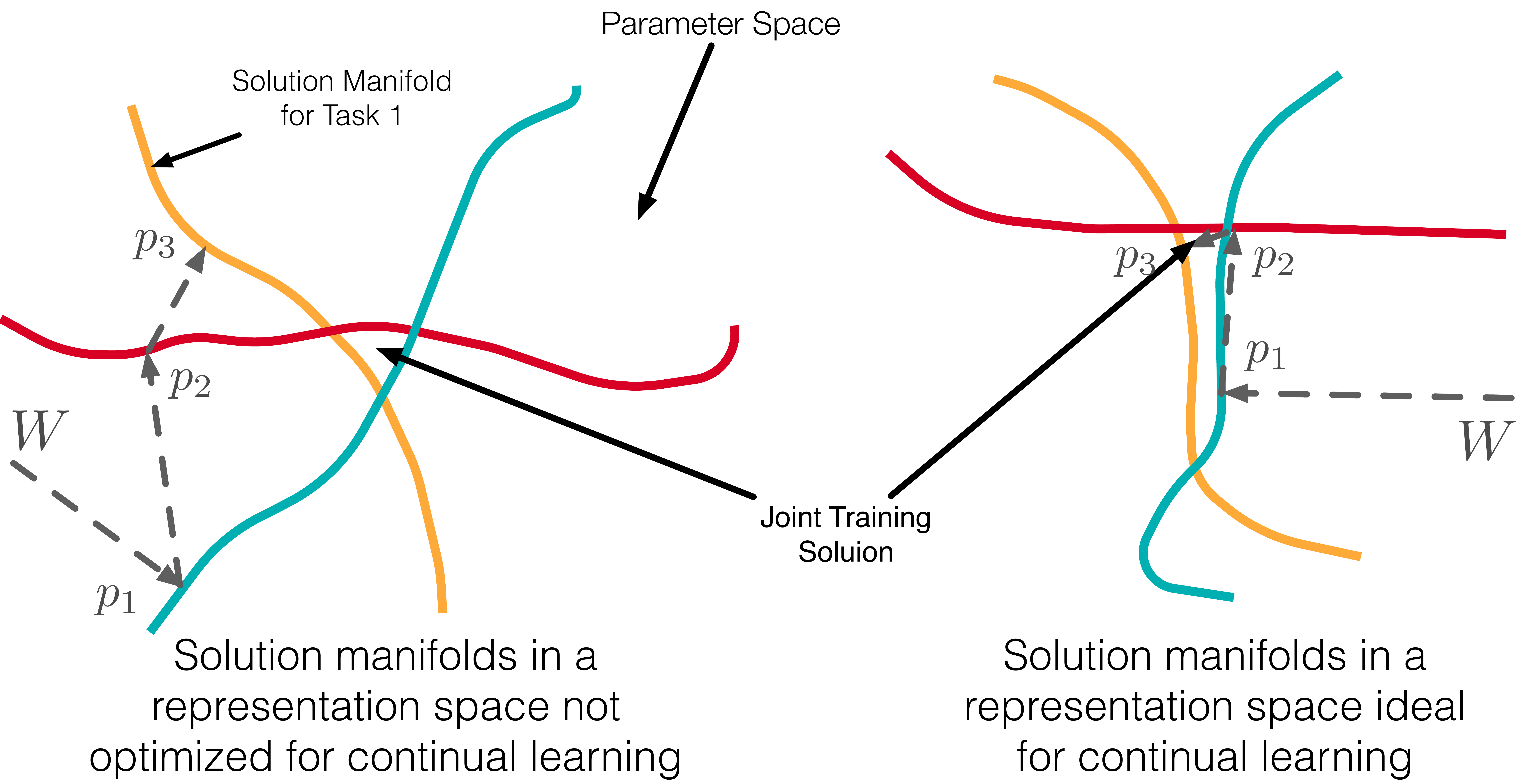}
	\caption{Effect of the representation on continual learning, for a problem where targets are generated from three different distributions $p_1(Y | x), p_2(Y|x)$ and $p_3(Y|x)$. The representation results in different solution manifolds for the three distributions; we depict two different possibilities here. 
		We show the learning trajectory when training incrementally from data generates first by $p_1$, then $p_2$ and $p_3$. On the left, the online updates interfere, jumping between distant points on the manifolds. On the right, the online updates either generalize appropriately---for parallel manifolds---or avoid interference because manifolds are orthogonal.
	}
	\label{fig:parameter_space}
\end{figure}

\section{Meta-learning Representations for Continual Learning}
Neural networks, trained end-to-end, are not effective at minimizing the $\CLP$  loss using a single trajectory sampled from $p(\mathcal S_k|\task)$ for two reasons. First, they are extremely sample-inefficient,  requiring multiple epochs of training to converge to reasonable solutions. Second, they suffer from catastrophic interference when learning online from a correlated stream of data \citep{french1991using}. Meta-learning is effective at making neural networks more sample efficient \citep{finn2017model}. Recently, \citet{nagabandi2018deep, al2017continuous} showed that it can also be used for quick adaptation from a stream of data. However, they do not look at the catastrophic interference problem. Moreover, their work meta-learns a model initialization, an inductive bias we found insufficient for solving the catastrophic interference problem (See Appendix~\ref{initvsrep}).

To apply neural network to the CLP problem, we propose meta-learning a function $\phi_{\repparams}(\inp)$ -- a deep Representation Learning Network (RLN) parametrized by $\repparams$ -- from $\mathcal{X}  \rightarrow  \mathbb{R}^\repdim$. We then learn another function $g_{\taskparams}$ from $\mathbb{R}^\repdim \rightarrow  \mathcal{Y}$, called a Prediction Learning Network (PLN). By composing the two functions we get $f_{\taskparams, \repparams}(\inp) = g_\taskparams(\phi_\repparams(\inp))$, which constitute our model for the CLP tasks as shown in Figure~\ref{fig:overview}.  We treat $\repparams$ as meta-parameters that are learned by minimizing a meta-objective and then later fixed at meta-test time. After learning $\repparams$, we learn $g_{\taskparams}$ from $\mathbb{R}^\repdim \rightarrow  \mathcal{Y}$ for a CLP problem from a single trajectory $\mathcal S$ using fully online SGD updates in a single pass. A similar idea has been proposed by \citet{bengio2019meta} for learning causal structures. 

For meta-training, we assume a distribution over CLP problems given by $p(\task)$. We consider two meta-objectives for updating the meta-parameters $\repparams$. (1) MAML-Rep, a MAML \citep{finn2017model} like few-shot-learning objective that learns an RLN instead of model initialization, and $\OML$ (Online aware Meta-learning) -- an objective that also minimizes interference in addition to maximizing fast adaptation for learning the RLN. Our $\OML$ objective is defined as: 

\begin{equation}
	\label{obj_oml}
	 \min_{\taskparams,\repparams} \sum_{\task_i \sim p(\task)}  \OML(\taskparams, \repparams) \defeq \sum_{\task_i \sim p(\task)}  \sum_{\mathcal S_k^j \sim p(\mathcal S_k | \task_i)}\left[\mathcal{L}_{CLP_i} \Big(U(\taskparams,\repparams,  \mathcal S_k^j) \right] 
	\end{equation}	

where $S^j_k=(X^i_{j+1} Y^i_{j+1}), (X^i_{j+2} Y^i_{j+2}), \dots, (X^i_{j+k} Y^i_{j+k})$. $U( \taskparams_t,\repparams, S^j_k) = (\taskparams_{t+\numupdates}, \repparams$) represents an update function where $\taskparams_{t+\numupdates}$ is the weight vector after $\numupdates$ steps of stochastic gradient descent. The $jth$ update step in $U$ is taken using parameters $(\taskparams_{t+j-1}, \repparams)$ on sample $(X^i_{t+j}, Y^i_{t+j})$ to give $(\taskparams_{t+j}, \repparams)$. 

MAML-Rep and $\OML$ objectives can be implemented as Algorithm~\ref{MAMLalgorithm} and \ref{algorithm} respectively, with the primary difference between the two highlighted in blue. Note that MAML-Rep uses the complete batch of data $\mathcal S_k$ to do $l$ inner updates (where $l$ is a hyper-parameter) whereas $\OML$ uses one data point from $\mathcal S_{k}$ for one update. This allows $\OML$ to take the effects of online continual learning -- such as catastrophic forgetting -- into account.

% Meta-parameters are learned during the meta-training phase and then fixed at meta-testing. They can be used to learn several aspects of the learning process. For instance, we can learn a model initialization \citep{finn2017model,li2017meta}, an update rule \citep{metz2019meta, bengio1990learning}, a context vector \citep{zintgraf2019fast}, or an encoder \citep{bengio2019meta} (among others). 

\begin{wrapfigure}{R}{0.70\textwidth}
	\vspace{-20pt}
	\begin{minipage}{0.70\textwidth}
		\begin{algorithm}[H]
			\centering
			\caption{Meta-Training : MAML-Rep}\label{MAMLalgorithm}
			\begin{algorithmic}[1]
				\REQUIRE $p(\task)$: distribution over CLP problems
				\REQUIRE $\alpha$, $\beta$: step size hyperparameters
					\REQUIRE $l$: No of inner gradient steps
				\STATE randomly initialize $\repparams$
				\WHILE{not done}
				\STATE randomly initialize $\taskparams$
				\STATE Sample CLP problem $\task_i \sim p(\task)$
				
				\STATE Sample $\mathcal S_{train}$ from $p( \mathcal S_{k} | \task_i)$
				\STATE $\taskparams_0 = \taskparams$
				\color{blue} \FOR {$j$ in $1, 2, \dots , l$}
				\STATE   \color{blue} $\taskparams_j=\taskparams_{j-1}-\alpha
				\nabla_{\taskparams_{j-1}}   \loss_i (f_{\repparams, \taskparams_l}(S_{train}[:, 0]), S_{train}[:, 1])$ 
				\ENDFOR
				\color{black}
				\STATE Sample $S_{test}$ from $p( \mathcal S_{k} | \task_i)$
				
				\STATE Update $\repparams \leftarrow \repparams - \beta \nabla_\repparams \loss_i (f_{\repparams, \taskparams_l}(S_{test}[:, 0]), S_{test}[:, 1])$ 
				\ENDWHILE
			\end{algorithmic}
		\end{algorithm}
	\end{minipage}
	\vspace{-10pt}
\end{wrapfigure}

The goal of the $\OML$ objective is to learn representations suitable for online continual learnings. For an illustration of what would constitute an effective representation for continual learning, suppose that we have three clusters of inputs, which have significantly different $p(Y | x)$, corresponding to $p_1$, $p_2$ and $p_3$. For a fixed 2-dimensional representation  $\phi_\repparams: \mathcal{X} \rightarrow \mathbb{R}^2$, we can consider the manifold of solutions $\taskparams \in \mathbb{R}^2$ given by a linear model that provide equivalently accurate solutions for each $p_i$.  These three manifolds are depicted as three different colored lines in the $\taskparams \in \mathbb{R}^2$  parameter space in Figure \ref{fig:parameter_space}. 
The goal is to find one parameter vector $\taskparams$ that is effective for all three distributions by learning online on samples from three distributions sequentially. For two different representations, these manifolds, and their intersections can look very different. The intuition is that online updates from a $\taskparams$ are more effective when the manifolds are either parallel---allowing for positive generalization---or orthogonal---avoiding interference. It is unlikely that a representation producing such manifolds would emerge naturally. Instead, we will have to explicitly find it. By taking into account the effects of online continual learning, the $\OML$ objective optimizes for such a representation.

We can optimize this objective similarly to other gradient-based meta-learning objectives. Early work on learning-to-learn considered optimizing parameters through learning updates themselves, though typically considering approaches using genetic algorithms \citep{schmidhuber1987evolutionary}. Improvements in automatic differentiation have made it more feasible to compute gradient-based meta-learning updates \citep{Finn:EECS-2018-105}. Some meta-learning algorithms have similarly considered optimizations through multiple steps of updating for the few-shot learning setting \citep{finn2017model, li2017meta, al2017continuous, nagabandi2018deep} for learning model initializations. The successes in these previous works in optimizing similar objectives motivate $\OML$ as a feasible objective for Meta-learning Representations for Continual Learning.

\begin{wrapfigure}{R}{0.62\textwidth}
	\vspace{-25pt}
	\begin{minipage}{0.62\textwidth}
		\begin{algorithm}[H]
			\centering
			\caption{Meta-Training : $\OML$}\label{algorithm}
			\begin{algorithmic}[1]
				\REQUIRE $p(\task)$: distribution over CLP problems
				\REQUIRE $\alpha$, $\beta$: step size hyperparameters
				\STATE randomly initialize $\repparams$
				\WHILE{not done}
				\STATE randomly initialize $\taskparams$
				\STATE Sample CLP problem $\task_i \sim p(\task)$
				\STATE Sample $\mathcal S_{train}$ from $p( \mathcal S_{k} | \task_i)$
				\STATE $\taskparams_0 = \taskparams$
				\color{blue}
				\FOR {$j = 1, 2, \dots, k$}
				\STATE $(X_j, Y_j) = \mathcal S_{train}[j]$
				\STATE  $\taskparams_j=\taskparams_{j-1}-\alpha
				\nabla_{\taskparams_{j-1}}  \loss_i(f_{\repparams, \taskparams_{j-1}}(X_{j}), Y_j)$ 
				\ENDFOR
				\color{black}
				\STATE Sample $S_{test}$ from $p( \mathcal S_{k} | \task_i)$
				
				\STATE Update $\repparams \leftarrow \repparams - \beta \nabla_\repparams \loss_i (f_{\repparams, \taskparams_{k}}(S_{test}[:, 0]), S_{test}[:, 1])$

				\ENDWHILE
			\end{algorithmic}
		\end{algorithm}
	\end{minipage}
	\vspace{-10pt}

\end{wrapfigure}
% \begin{figure}
% 	\begin{algorithm}[H]
% 		\SetAlgoLined
% 		\textbf{Require:} $\mathcal D_{stream} = (X_1, Y_1), (X_2, Y_2), \ldots, (X_t, Y_t),
% 		\ldots$\;
% 		\textbf{Require:} $g_\taskparams(\phi_\repparams(x))$ as a parametrized function\;
% 		\textbf{Require:} $\alpha, \beta, \mathcal L, n$ as meta learning rate, inner learning rate, loss metric over $(X_i, Y_i)$ and total gradient updates\;
% 		%		\KwRequire{Algorithm for Learning Prior}
% 		Initialize RLN and TLN to $\theta$ and $W$\;
% 		\For{iterations $1,2,3,\dots,n$}{
% 			Sample trajectory $ (\mathbf{X_{traj}}, \mathbf{Y_{traj}}) = (X_{i+1}, Y_{i+1})\ldots(X_{i+k}, Y_{i+k}) \sim \mathcal D_{stream}$\;
% 			$W_0=W$\;
% 			\For{$j$ in $1,2,3,\ldots,k$}{
% 				$W_j =  W_{j-1} - \beta \nabla_{W_{j-1}}(\mathcal L(g_{W_{j-1}}(\phi_\repparams(X_{i+j})), Y_{i+j})$\;
% 			}
% 			$ (\mathbf{X_{rand}}, \mathbf{Y_{rand}}) \sim \mathcal D_{stream}$;
% 			\hfill Sample a random batch of data	\\
% 			$(\mathbf{X_{meta}}, \mathbf{Y_{meta}}) =(\mathbf{X_{rand}}+\mathbf{X_{traj}}, \mathbf{Y_{rand}}+\mathbf{Y_{traj}})$\;
% 			$W, \theta =(W, \theta) - \alpha \mathcal r_{W, \theta}\mathcal
% 			L(g_{W_{k}}(\phi_\repparams(\mathbf{X_{meta}})), \mathbf{Y_{meta}})$\;
% 		}
% 		\caption{\OML\ for optimizing objective in \eqref{obj_oml}} 
% 		\label{algo:reptile}
% 	\end{algorithm}
% \end{figure}

\section{Evaluation}

In this section, we investigate the question: can we learn a representation for continual learning that promotes future learning and reduces interference? We investigate this question by meta-learning the representations offline on a meta-training dataset. At meta-test time, we initialize the continual learner with this representation and measure prediction error as the agent learns the PLN online on a new set of CLP problems (See Figure~\ref{fig:overview}).

\subsection{CLP Benchmarks}
We evaluate on a simulated regression problem and a sequential classification problem using real data. 

\textbf{Incremental Sine Waves:} An Incremental Sine Wave CLP problem is defined by ten (randomly generated) sine functions, with $x = (z, n)$ for $z \in [-5, 5]$ as input to the sine function and $n$ a one-hot vector for $\{1, \ldots, 10\}$ indicating which function to use. The targets are deterministic, where $(x, y)$ corresponds to $y = \text{sin}_n(z)$. Each sine function is generated once by randomly selecting an amplitude in the range $[0.1,5]$ and phase in $[0, \pi]$. A trajectory $\mathcal S_{400}$ from the CLP problem consists of 40 mini-batches from the first sine function in the sequence (Each mini-batch has eight elements), and then 40 from the second and so on. Such a trajectory has sufficient information to minimize loss for the complete CLP problem. 
We use a single regression head to predict all ten functions, where the input id $n$ makes it possible to differentiate outputs for the different functions. Though learnable, this input results in significant interference across different functions.   

\textbf{Split-Omniglot:} 
% Omniglot is a dataset of over 1623 characters from 50 different alphabets \citep{lake2015human}. Each character has 20 hand-written images. The dataset is divided into two parts. The first 963 classes are the background images, which are used for learning a fixed representation, and the rest are for evaluation. For benchmarking, we learn online on the evaluation classes such that we see all sample of one class before going to the next. Moreover, we use the “single-pass through the data” protocol used by \citet{lopez2017gradient}. Our benchmark is similar to Split-Omniglot used by \citet{riemer2018learning} with each task consisting of a single class. Note that we make no use of task IDs at any time. To gauge both forgetting and generalization, we further split this evaluation set into training and test. We use 15 images for online training and 5 to compute test error.
Omniglot is a dataset of over 1623 characters from 50 different alphabets \citep{lake2015human}. Each character has 20 hand-written images. The dataset is divided into two parts. The first 963 classes constitute the meta-training dataset whereas the remaining 660 the meta-testing dataset. To define a CLP problem on this dataset, we sample an ordered set of 200 classes $(C_{1},C_{2},C_{3}, \dots, C_{200})$. $\mathcal X$ and $\mathcal Y$, then, constitute of all images of these classes. A trajectory $\mathcal S_{1000}$ from such a problem is a trajectory of images -- five images per class -- where we see all five images of $C_1$ followed by five images of $C_2$ and so on. This makes $k= 5 \times 200 = 1000$. Note that the sampling operation defines a distribution $p(\task)$ over problems that we use for meta-training.

\subsection{Meta-Training Details}
\begin{figure}
	\centering
 	\includegraphics[width=0.99\linewidth]{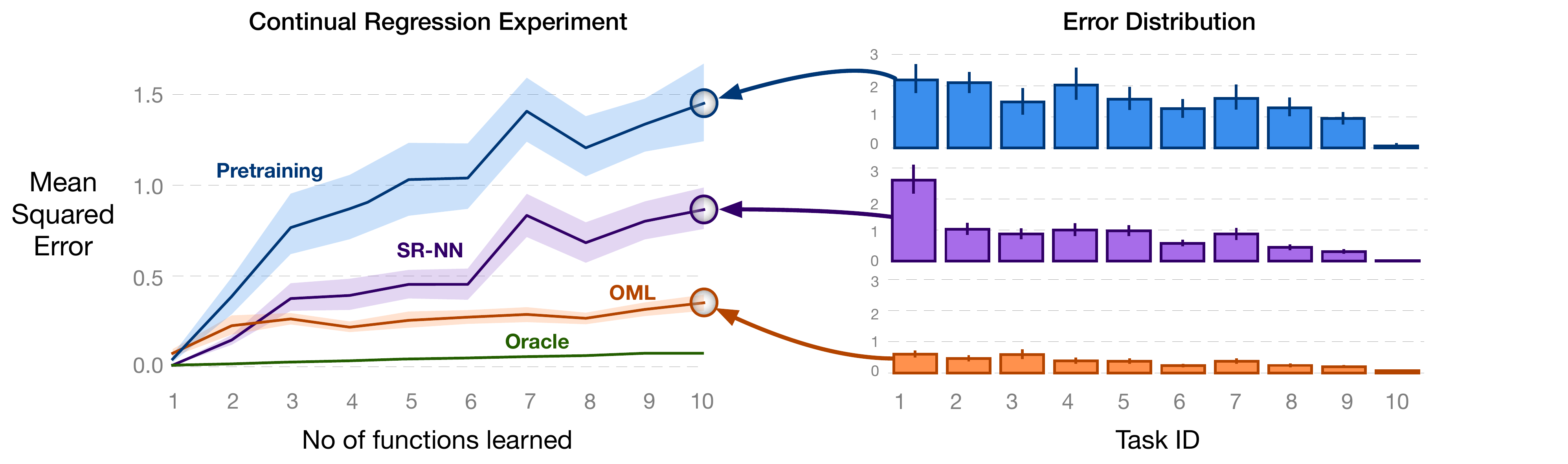}
	\caption{Mean squared error across all 10 regression tasks. The x-axis in (a) corresponds to seeing all data points of samples for class 1, then class 2 and so on. These learning curves are averaged over 50 runs, with error bars representing 95\% confidence interval drawn by 1,000 bootstraps. We can see that the representation trained on iid data---Pre-training---is not effective for online updating. Notice that in the final prediction accuracy in (b), Pre-training and SR-NN representations have accurate predictions for task 10, but high error for earlier tasks. $\OML$, on the other hand, has a slight skew in error towards later tasks in learning but is largely robust. Oracle uses iid sampling and multiple epochs and serves as a best case bound.}
	\label{fig:regression}
\end{figure}

\textbf{Incremental Sine Waves:}
We sample 400 functions to create our meta-training set and 500 for benchmarking the learned representation. We meta-train by sampling multiple CLP problems. During each meta-training step, we sample ten functions from our meta-training set and assign them task ids from one to ten. We concatenate 40 mini-batches generated from function one, then function two and so on, to create our training trajectory $\mathcal S_{400}$. For evaluation, we similarly randomly sample ten functions from the test set and create a single trajectory. 
We use SGD on the MSE loss with a mini-batch size of 8 for online updates,  and Adam \citep{kingma2014adam}  for optimizing the $\OML$ objective. Note that the $\OML$ objective involves computing gradients through a network unrolled for 400 steps. 
At evaluation time, we use the same learning rate as used during the inner updates in the meta-training phase for \OML. For our baselines, we do a grid search over learning rates and report the results for the best performing parameter. 

We found that having a deeper representation learning network (RLN) improved performance.  We use six layers for the RLN and two layers for the PLN. Each hidden layer has a width of 300. The RLN is only updated with the meta-update and acts as a fixed feature extractor during the inner updates in the meta-learning objective and at evaluation time. 
 
 \textbf{Split-Omniglot:}
 We learn an encoder -- a deep CNN with 6 convolution and two FC layers -- using the MAML-Rep and the $\OML$ objective. We treat the convolution parameters as $\repparams$ and FC layer parameters as $\taskparams$. Because optimizing the $\OML$ objective is computationally expensive for $H=1000$ (It involves unrolling the computation graph for 1,000 steps), we approximate the two objectives. For MAML-Rep we learn the $\phi_\repparams$ by maximizing fast adaptation for a 5 shot 5-way classifier. For $\OML$, instead of doing $|\mathcal S_{train}|$ no of inner-gradient steps as described in Algorithm~\ref{algorithm}, we go over $\mathcal S_{train}$ five steps at a time. For $kth$ five steps in the inner loop, we accumulate our meta-loss on $\mathcal S_{test}[0:5\times k]$, and update our meta-parameters using these accumulated gradients at the end as explained in Algorithm \ref{algorithm_approx} in the Appendix.  This allows us to never unroll our computation graphs for more than five steps (Similar to truncated back-propagation through time) and still take into account the effects of interference at meta-training.

Finally, both MAML-Rep and $\OML$ use 5 inner gradient steps and similar network architectures for a fair comparison. Moreover, for both methods, we try multiple values for the inner learning rate $\alpha$ and report the results for the best parameter. For more details about hyper-parameters see the Appendix.  For more details on implementation, see Appendix \ref{app_exp}.

%  The experimental details are similar, except we use SGD on the cross-entropy loss and a six-layer convolutional neural network for the RLN. 
 
% For the meta-update, it is not computationally feasible to compute gradients through the update for the complete trajectory of 1000. Instead, we (a) sample a random batch of data from the meta-training set; (b) we sample a single class and then (c) compute the gradient of the loss for this random batch through an online update for the data for that single class (One gradient step for each of the 20 examples). 

\subsection{Baselines}
We compare MAML-Rep and $\OML$ -- the two Meta-learneing based Representations Leanring methods to three baselines. 

\textbf{Scratch} simply learns online from a random network initialization, with no meta-training. 

\textbf{Pre-training} uses standard gradient descent to minimize prediction error on the meta-training set. We then fix the first few layers in online training. Rather than restricting to the same 6-2 architecture for the RLN and PLN, we pick the best split using a validation set. 

\textbf{SR-NN} use the Set-KL method to learn a sparse representation \citep{liu2018utility} on the meta-training set. We use multiple values of the hyper-parameter $\beta$ for SR-NN and report results for one that performs the best. We include this baseline to compare to a method that learns a sparse representation.

 \subsection{Meta-Testing}
 
 \begin{figure}
 	\centering
 	\includegraphics[width=0.99\linewidth]{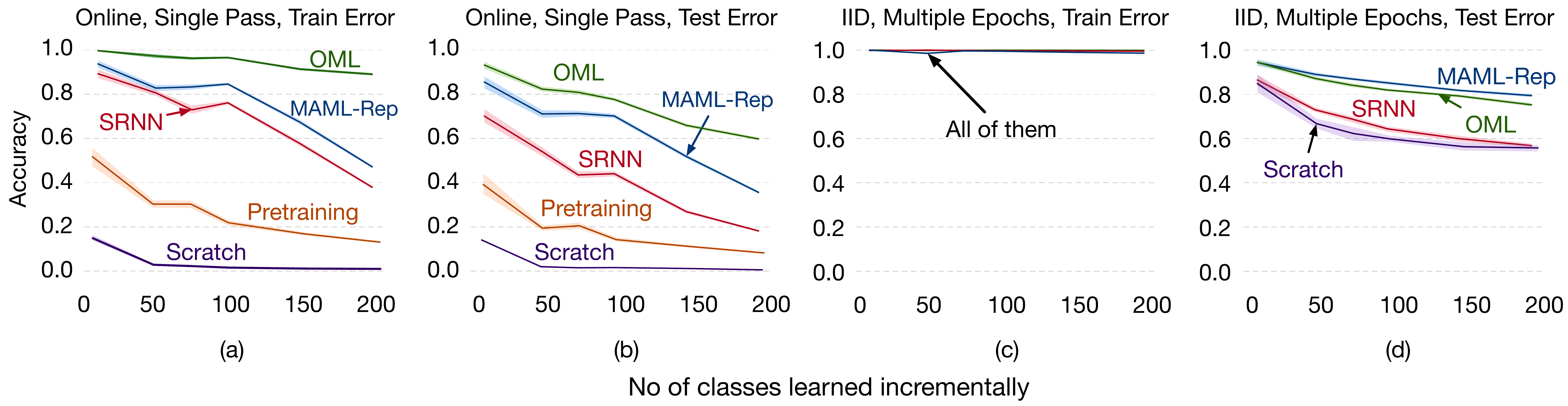}
	 \caption{Comparison of representations learned by the MAML-Rep, $\OML$ objective and the baselines on Split-Omniglot. All curves are averaged over 50 CLP runs with 95\% confidence intervals drawn using 1,000 bootstraps. At every point on the x-axis, we only report accuracy on the classes seen so far. Even though both MAML-Rep and $\OML$ learn representations that result in comparable performance of classifiers trained under the IID setting (c and d), $\OML$ out-performs MAML-Rep when learning online on a highly correlated stream of data showing it learns representations more robust to interference. SR-NN, which does not do meta-learning, performs worse even under the IID setting showing it learns worse representations.}
 	\label{fig:classification}
 \end{figure}

 We report results of $\mathcal L_{CLP}(\taskparams_{online}, \repparams_{meta})$ for fully online updates on a single $\mathcal S_k$ for each CLP problem. For each of the methods, we separately tune the learning rate on a five validation trajectories and report results for the best performing parameter.
 
\textbf{Incremental Sine Waves:} 
We plot the average mean squared error over 50 runs on the full testing set, when learning online on unseen sequences of functions, in Figure~\ref{fig:regression}~(left). \OML\ can learn new functions with a negligible increase in average MSE. The Pre-training baseline, on the other hand, clearly suffers from interference, with increasing error as it tries to learn more and more functions. SR-NN, with its sparse representation, also suffers from noticeably more interference than \OML.   From the distribution of errors for each method on the ten functions, shown in Figure~\ref{fig:regression}~(right), we can see that both Pre-training and SR-NN have high errors for functions learned in the beginning whereas \OML\ performs only slightly worse on those.
 
\begin{wrapfigure}{R}{0.50\textwidth}

	\begin{center}
		\includegraphics[width=0.50\textwidth]{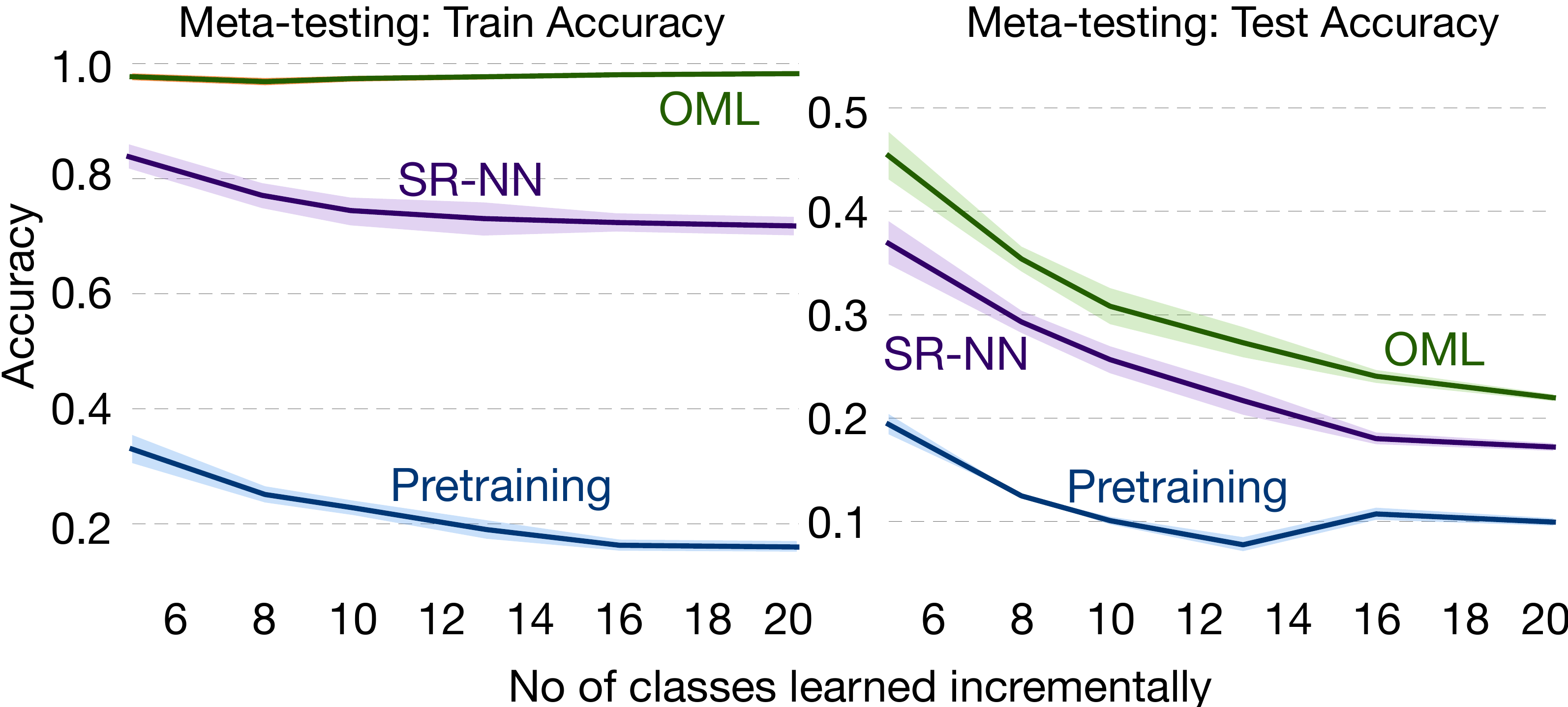}
	\end{center}
	\caption{$\OML$ scales to more complex datasets such a Mini-imagenet. We use the existing meta-training/meta-testing split of mini-imagenet. At meta-testing, we learn a 20 way classifier using 30 samples per class.}
	\label{imagenet}
	\vspace{-20pt}
\end{wrapfigure}

 \textbf{Split-Omniglot:}
 
 We report classification accuracy on the training trajectory ($\mathcal S_{train}$) as well as the test set in Figure~\ref{fig:classification}. Note that training accuracy is a meaningful metric in continual learning as it measures forgetting.  The test set accuracy reflects both forgetting and generalization error. Our method can learn the training trajectory almost perfectly with minimal forgetting. The baselines, on the other hand, suffer from forgetting as they learn more classes sequentially. The higher training accuracy of our method also translates into better generalization on the test set. The difference in the train and test performance is mainly due to how few samples are given per class: only 15 for training and 5 for testing. 

 As a sanity check, we also trained classifiers by sampling data IID for 5 epochs and report the results in Fig. \ref{fig:classification} (c) and (d). The fact that \OML\ and MAML-Rep do equally well with IID sampling indicates that the quality of representations ($\phi_{\repparams} = \mathbb{R}^\repdim$) learned by both objectives are comparable and the higher performance of \OML\ is indeed because the representations are more suitable for incremental learning. 
 
 Moreover, to test if $\OML$ can learn representations on more complex datasets, we run the same experiments on mini-imagenet and report the results in Figure~\ref{imagenet}.

\begin{figure}
	\centering
	\includegraphics[width=0.99\linewidth]{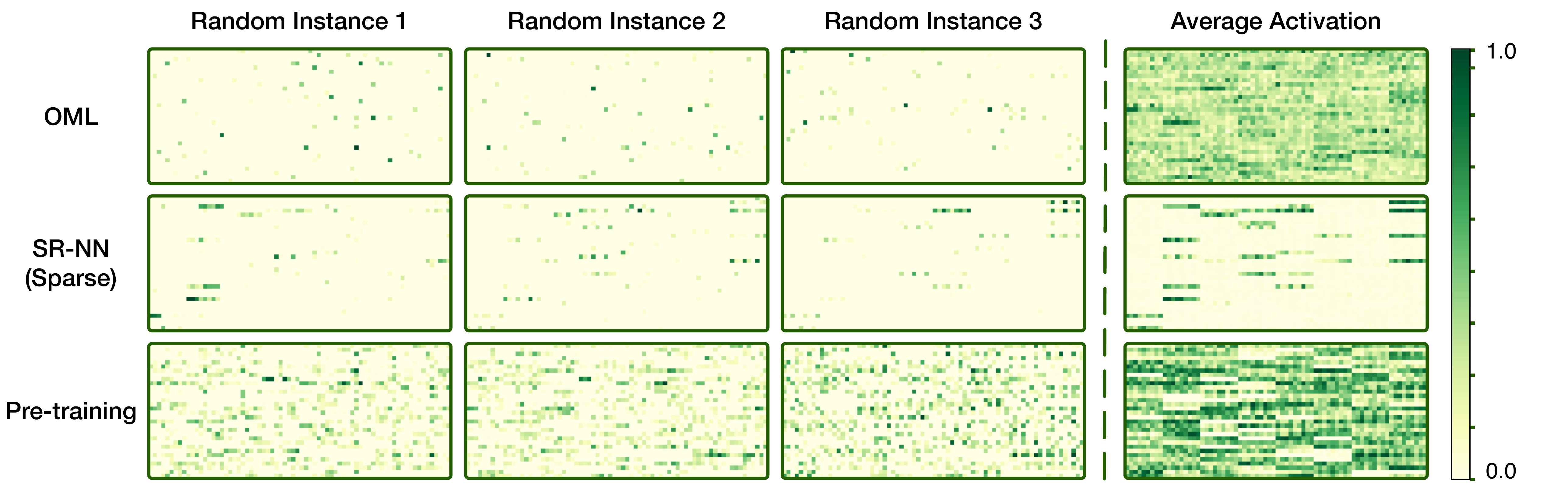}
	\caption{We reshape the 2304 length representation vectors into 32x72, normalize them to have a maximum value of one and visualize them; here random instance means representation for a randomly chosen input from the training set, whereas average activation is the mean representation for the complete dataset. For SR-NN, we re-train the network with a different value of parameter $\beta$ to have the same instance sparsity as $\OML$. Note that SR-NN achieves this sparsity by never using a big part of representation space. \OML, on the other hand, uses the full representation space. In-fact, \OML\ has no dead neurons whereas even pre-training results in some part of the representation never being used.}
	\label{fig:rep_comparison}
\end{figure}
	
\subsection{What kind of representations does \OML\ learn?}

As discussed earlier, \citet{french1991using} proposed that sparse representations could mitigate forgetting. Ideally, such a representation is instance sparse--using a small percentage of activations to represent an input-- while also utilizing the representation to its fullest. This means that while most neurons would be inactive for a given input, every neuron would participate in representing some input. Dead neurons, which are inactive for all inputs, are undesirable and may as well be discarded. An instance sparse representation with no dead neurons reduces forgetting because each update changes only a small number of weights which in turn should only affect a small number of inputs. 
We hypothesize that the representation learned by \OML\ will be sparse, even though the objective does not explicitly encourage this property. 

We compute the average instance sparsity on the Omniglot training set, for \OML, SR-NN, and Pre-training. \OML\ produces the most sparse network, without any dead neurons. The network learned by Pre-training, in comparison, uses over 10 times more neurons on average to represent an input. The best performing SR-NN used in Figure \ref{fig:classification} uses 4 times more neurons. We also re-trained SR-NN with a parameter to achieve a similar level of sparsity as \OML, to compare representations of similar sparsity rather than representations chosen based on accuracy. We use $\beta=0.05$ which results in an instance sparsity similar to \OML. 

\begin{wraptable}{R}{0.55\textwidth}
	\vspace{-20pt}
	\centering
	\caption{Instance sparisty and dead neuron percentage for different methods. $\OML$ learns highly sparse representations without any dead neurons. Even Pre-training, which does not optimize for sparsity, ends up with some dead neurons, on the other hand.}
	\begin{minipage}{0.55\textwidth}
		% \begin{table}

			\begin{tabular}{ccc}
				\toprule
				\centering
				Method &   Instance Sparsity & Dead Neurons  \\
				\midrule
				$\OML$ & 3.8\% & 0\% \\ 
				SR-NN (Best) & 15\%  & 0.7\% \\ 
				SR-NN (Sparse) & 4.9\%  & 14\% \\ 
				Pre-Training & 38\%  & 3\%  \\ 
				\bottomrule
			\end{tabular}
		% \end{table}
	\end{minipage}
	\label{sparsity_table}
	\vspace{-10pt}

\end{wraptable}
	
We visualize all the solutions in Figure \ref{fig:rep_comparison}. The plots highlight that \OML\ learns a highly sparse and well-distributed representation, taking the most advantage of the large capacity of the representation. Surprisingly, \OML\ has no dead neurons, which is a well-known problem when learning sparse representations \citep{liu2018utility}. Even Pre-training, which does not have an explicit penalty to enforce sparsity, has some dead neurons. Instance sparsity and dead neurons percentage for each method are reported in Table~\ref{sparsity_table}.

\section{Improvements by Combining with Knowledge Retention Approaches}
We have shown that \OML\ learns effective representations for continual learning. In this section, we answer a different question: how does \OML\ behave when it is combined with existing continual learning methods? We test the performance of EWC \citep{lee2017overcoming}, MER \citep{riemer2018learning} and ER-Reservoir \citep{chaudhry2019continual}, in their standard form---learning the whole network online---as well as with pre-trained fixed representations. We use pre-trained representations from \OML\ and Pre-training, obtained in the same way as described in earlier sections. For the Standard online form of these algorithms, to avoid the unfair advantage of meta-training, we initialize the networks by learning iid on the meta-training set.  %We also allow the continual learning methods to change the whole network In the third case, we use the same network as Pre-training network, except we do not fix the initial RLN layers. Consequently, this network learns directly from the raw observation space. 

As baselines, we also report results for (a) fully online SGD updates that update one point at a time in order on the trajectory and (b) approximate IID training where SGD updates are used on a random shuffling of the trajectory, removing the correlation. 

\begin{table}
	\caption{\OML\ combined with existing continual learning methods. All memory-based methods use a buffer of 200. Error margins represent one std over 10 runs. Performance of all methods is considerably improved when they learn from representations learned by \OML\; moreover, even online updates are competitive with rehearsal based methods with $\OML$. Finally, online updates on \OML\ outperform all methods when they learn from other representations. Note that MER does better than approx IID in some cases because it does multiple rehearsal-based updates for every sample.}
	\label{comparison}
	\centering
	\begin{tabular}{lllllll}
		\toprule
		&    \multicolumn{6}{c}{Split-Omniglot}       \\
		\cmidrule(r){2-7} 
		& \multicolumn{3}{c}{One class per task, 50 tasks} & \multicolumn{3}{c}{Five classes per task, 20 tasks}\\
		\cmidrule(r){2-4}  	\cmidrule(r){5-7} 
		Method &   Standard & \OML\ & Pre-training &   Standard & \OML\ & Pre-training   \\
		\midrule
		Online &04.64  {\tiny$\pm$ 2.61}  & \textbf{64.72}   {\tiny$\pm$ 2.57 }  & 21.16   {\tiny$\pm$ 2.71} & 01.40  {\tiny$\pm$ 0.43}  & \textbf{55.32} {\tiny$\pm$ 2.25}   & 11.80 {\tiny$\pm$ 1.92} \\ 
		Approx IID &53.95   {\tiny$\pm$ 5.50} & \textbf{75.12}  {\tiny$\pm$ 3.24}  & 54.29 {\tiny$\pm$ 3.48} & 48.02 {\tiny$\pm$ 5.67} & \textbf{67.03} {\tiny$\pm$ 2.10}& 46.02 {\tiny$\pm$ 2.83}  \\
		ER-Reservoir & 52.56  {\tiny$\pm$ 2.12}  & \textbf{68.16}   {\tiny$\pm$ 3.12} & 36.72 {\tiny$\pm$ 3.06} & 24.32  {\tiny$\pm$ 5.37} & \textbf{60.92}  {\tiny$\pm$ 2.41} & 37.44 {\tiny$\pm$ 1.67}  \\
		MER & 54.88  {\tiny$\pm$ 4.12} & \textbf{76.00} {\tiny$\pm$ 2.07}    & 62.76 {\tiny$\pm$ 2.16} & 29.02 {\tiny$\pm$ 4.01} &\textbf{62.05} {\tiny$\pm$ 2.19} & 42.05  {\tiny$\pm$ 3.71}  \\
		EWC & 05.08 {\tiny$\pm$ 2.47} &\textbf{64.44} {\tiny$\pm$ 3.13}& 18.72 {\tiny$\pm$ 3.97}& 02.04 {\tiny$\pm$ 0.35} & \textbf{56.03} {\tiny$\pm$ 3.20} & 10.03 {\tiny$\pm$ 1.53} \\
		
		\bottomrule
	\end{tabular}
\end{table}
We report the test set results for learning 50 tasks with one class per task and learning 20 tasks with 5 tasks per class in Split-Omniglot in Table \ref{comparison}. For each of the methods, we do a 15/5 train/test split for each Omniglot class and test multiple values for all the hyperparameters and report results for the best setting. The conclusions are surprisingly clear. (1) \OML\ improves all the algorithms; (2) simply providing a fixed representation, as in Pre-training, does not provide nearly the same gains as \OML\; and (3) \OML\ with a basic Online updating strategy is already competitive, outperforming all the continual learning methods without \OML. 
There are a few additional outcomes of note. \OML\ outperforms even approximate IID sampling, suggesting it is not only mitigating interference but also making learning faster on new data. Finally, the difference in online and experience replay based algorithms for \OML\ is not as pronounced as it is for other representations. 

\section{Conclusion and Discussion} 

In this paper, we proposed a meta-learning objective to learn representations that are robust to interference under online updates and promote future learning. We showed that using our representations, it is possible to learn from highly correlated data streams with significantly improved robustness to forgetting. We found sparsity emerges as a property of our learned representations, without explicitly training for sparsity. We finally showed that our method is complementary to the existing state of the art continual learning methods, and can be combined with them to achieve significant improvements over each approach alone.

An important next step for this work is to demonstrate how to learn these representations online without a separate meta-training phase. Initial experiments suggest it is effective to periodically optimize the representation on a recent buffer of data, and then continue online update with this updated fixed representation. This matches common paradigms in continual learning---based on the ideas of a sleep phase and background planning---and is a plausible strategy for continually adapting the representation network for a continual stream of data. Another interesting extension to the work would be to use the \OML\ objective to meta-learn some other aspect of the learning process -- such as a local learning rule \citep{metz2019meta} or an attention mechanism -- by minimizing interference.

	\bibliographystyle{chicagoo}

	\bibliography{noniid}

\begin{thebibliography}{}

\bibitem[\protect\citeauthoryear{Al-Shedivat, Bansal, Burda, Sutskever,
  Mordatch, and Abbeel}{Al-Shedivat et~al.}{2018}]{al2017continuous}
Al-Shedivat, Maruan, Trapit Bansal, Yuri Burda, Ilya Sutskever, Igor Mordatch,
  and Pieter Abbeel (2018).
\newblock Continuous adaptation via meta-learning in nonstationary and
  competitive environments.
\newblock {\em International Conference on Learning Representations\/}.

\bibitem[\protect\citeauthoryear{Aljundi, Babiloni, Elhoseiny, Rohrbach, and
  Tuytelaars}{Aljundi et~al.}{2018}]{aljundi2018memory}
Aljundi, Rahaf, Francesca Babiloni, Mohamed Elhoseiny, Marcus Rohrbach, and
  Tinne Tuytelaars (2018).
\newblock Memory aware synapses: Learning what (not) to forget.
\newblock In {\em European Conference on Computer Vision}.

\bibitem[\protect\citeauthoryear{Aljundi, Lin, Goujaud, and Bengio}{Aljundi
  et~al.}{2019}]{aljundi19}
Aljundi, Rahaf, Min Lin, Baptiste Goujaud, and Yoshua Bengio (2019).
\newblock Gradient based sample selection for online continual learning.
\newblock {\em Advances in Neural Information Processing Systems\/}.

\bibitem[\protect\citeauthoryear{Aljundi, Rohrbach, and Tuytelaars}{Aljundi
  et~al.}{2019}]{aljundi2018selfless}
Aljundi, Rahaf, Marcus Rohrbach, and Tinne Tuytelaars (2019).
\newblock Selfless sequential learning.
\newblock {\em International Conference on Learning Representations\/}.

\bibitem[\protect\citeauthoryear{Bengio, Deleu, Rahaman, Ke, Lachapelle,
  Bilaniuk, Goyal, and Pal}{Bengio et~al.}{2019}]{bengio2019meta}
Bengio, Yoshua, Tristan Deleu, Nasim Rahaman, Rosemary Ke, S{\'e}bastien
  Lachapelle, Olexa Bilaniuk, Anirudh Goyal, and Christopher Pal (2019).
\newblock A meta-transfer objective for learning to disentangle causal
  mechanisms.
\newblock {\em arXiv preprint arXiv:1901.10912\/}.

\bibitem[\protect\citeauthoryear{Chaudhry, Ranzato, Rohrbach, and
  Elhoseiny}{Chaudhry et~al.}{2019}]{chaudhry2018efficient}
Chaudhry, Arslan, Marc'Aurelio Ranzato, Marcus Rohrbach, and Mohamed Elhoseiny
  (2019).
\newblock Efficient lifelong learning with a-gem.
\newblock {\em International Conference on Learning Representations\/}.

\bibitem[\protect\citeauthoryear{Chaudhry, Rohrbach, Elhoseiny, Ajanthan,
  Dokania, Torr, and Ranzato}{Chaudhry et~al.}{2019}]{chaudhry2019continual}
Chaudhry, Arslan, Marcus Rohrbach, Mohamed Elhoseiny, Thalaiyasingam Ajanthan,
  Puneet~K Dokania, Philip~HS Torr, and Marc'Aurelio Ranzato (2019).
\newblock Continual learning with tiny episodic memories.
\newblock {\em arXiv:1902.10486\/}.

\bibitem[\protect\citeauthoryear{Finn}{Finn}{2018}]{Finn:EECS-2018-105}
Finn, Chelsea (2018, Aug).
\newblock {\em Learning to Learn with Gradients}.
\newblock Ph.\ D. thesis, EECS Department, University of California, Berkeley.

\bibitem[\protect\citeauthoryear{Finn, Abbeel, and Levine}{Finn
  et~al.}{2017}]{finn2017model}
Finn, Chelsea, Pieter Abbeel, and Sergey Levine (2017).
\newblock Model-agnostic meta-learning for fast adaptation of deep networks.
\newblock In {\em International Conference on Machine Learning}.

\bibitem[\protect\citeauthoryear{French}{French}{1991}]{french1991using}
French, Robert~M (1991).
\newblock Using semi-distributed representations to overcome catastrophic
  forgetting in connectionist networks.
\newblock In {\em Annual cognitive science society conference}. Erlbaum.

\bibitem[\protect\citeauthoryear{French}{French}{1999}]{french1999catastrophic}
French, Robert~M (1999).
\newblock Catastrophic forgetting in connectionist networks.
\newblock {\em Trends in cognitive sciences\/}.

\bibitem[\protect\citeauthoryear{Kingma and Ba}{Kingma and
  Ba}{2014}]{kingma2014adam}
Kingma, Diederik~P and Jimmy Ba (2014).
\newblock Adam: A method for stochastic optimization.
\newblock {\em arXiv:1412.6980\/}.

\bibitem[\protect\citeauthoryear{Kirkpatrick, Pascanu, Rabinowitz, Veness,
  Desjardins, Rusu, Milan, Quan, Ramalho, Grabska-Barwinska,
  et~al.}{Kirkpatrick et~al.}{2017}]{kirkpatrick2017overcoming}
Kirkpatrick, James, Razvan Pascanu, Neil Rabinowitz, Joel Veness, Guillaume
  Desjardins, Andrei~A Rusu, Kieran Milan, John Quan, Tiago Ramalho, Agnieszka
  Grabska-Barwinska, et~al. (2017).
\newblock Overcoming catastrophic forgetting in neural networks.
\newblock {\em National academy of sciences\/}.

\bibitem[\protect\citeauthoryear{Lake, Salakhutdinov, and Tenenbaum}{Lake
  et~al.}{2015}]{lake2015human}
Lake, Brenden~M, Ruslan Salakhutdinov, and Joshua~B Tenenbaum (2015).
\newblock Human-level concept learning through probabilistic program induction.
\newblock {\em Science\/}.

\bibitem[\protect\citeauthoryear{Lee, Kim, Jun, Ha, and Zhang}{Lee
  et~al.}{2017}]{lee2017overcoming}
Lee, Sang-Woo, Jin-Hwa Kim, Jaehyun Jun, Jung-Woo Ha, and Byoung-Tak Zhang
  (2017).
\newblock Overcoming catastrophic forgetting by incremental moment matching.
\newblock In {\em Advances in Neural Information Processing Systems}.

\bibitem[\protect\citeauthoryear{Li and Hoiem}{Li and
  Hoiem}{2018}]{li2018learning}
Li, Zhizhong and Derek Hoiem (2018).
\newblock Learning without forgetting.
\newblock {\em IEEE Transactions on Pattern Analysis and Machine
  Intelligence\/}.

\bibitem[\protect\citeauthoryear{Li, Zhou, Chen, and Li}{Li
  et~al.}{2017}]{li2017meta}
Li, Zhenguo, Fengwei Zhou, Fei Chen, and Hang Li (2017).
\newblock Meta-sgd: Learning to learn quickly for few-shot learning.
\newblock {\em arXiv:1707.09835\/}.

\bibitem[\protect\citeauthoryear{Lin}{Lin}{1992}]{lin1992self}
Lin, Long-Ji (1992).
\newblock Self-improving reactive agents based on reinforcement learning,
  planning and teaching.
\newblock {\em Machine learning\/}.

\bibitem[\protect\citeauthoryear{Liu, Kumaraswamy, Le, and White}{Liu
  et~al.}{2019}]{liu2018utility}
Liu, Vincent, Raksha Kumaraswamy, Lei Le, and Martha White (2019).
\newblock The utility of sparse representations for control in reinforcement
  learning.
\newblock {\em AAAI Conference on Artificial Intelligence\/}.

\bibitem[\protect\citeauthoryear{Liu, Masana, Herranz, Van~de Weijer, Lopez,
  and Bagdanov}{Liu et~al.}{2018}]{liu2018rotate}
Liu, Xialei, Marc Masana, Luis Herranz, Joost Van~de Weijer, Antonio~M Lopez,
  and Andrew~D Bagdanov (2018).
\newblock Rotate your networks: Better weight consolidation and less
  catastrophic forgetting.
\newblock In {\em International Conference on Pattern Recognition}.

\bibitem[\protect\citeauthoryear{Lopez-Paz and Ranzato}{Lopez-Paz and
  Ranzato}{2017}]{lopez2017gradient}
Lopez-Paz, David and Marc'Aurelio Ranzato (2017).
\newblock Gradient episodic memory for continual learning.
\newblock In {\em Advances in Neural Information Processing Systems}.

\bibitem[\protect\citeauthoryear{Metz, Maheswaranathan, Cheung, and
  Sohl-dickstein}{Metz et~al.}{2019}]{metz2019meta}
Metz, Luke, Niru Maheswaranathan, Brian Cheung, and Jascha Sohl-dickstein
  (2019).
\newblock Meta-learning update rules for unsupervised representation learning.
\newblock {\em International Conference on Learning Representations\/}.

\bibitem[\protect\citeauthoryear{Mnih, Kavukcuoglu, Silver, Rusu, Veness,
  Bellemare, Graves, Riedmiller, Fidjeland, Ostrovski, et~al.}{Mnih
  et~al.}{2015}]{mnih2015human}
Mnih, Volodymyr, Koray Kavukcuoglu, David Silver, Andrei~A Rusu, Joel Veness,
  Marc~G Bellemare, Alex Graves, Martin Riedmiller, Andreas~K Fidjeland, Georg
  Ostrovski, et~al. (2015).
\newblock Human-level control through deep reinforcement learning.
\newblock {\em Nature\/}.

\bibitem[\protect\citeauthoryear{Nagabandi, Finn, and Levine}{Nagabandi
  et~al.}{2019}]{nagabandi2018deep}
Nagabandi, Anusha, Chelsea Finn, and Sergey Levine (2019).
\newblock Deep online learning via meta-learning: Continual adaptation for
  model-based rl.
\newblock {\em International Conference on Learning Representations\/}.

\bibitem[\protect\citeauthoryear{Rebuffi, Kolesnikov, Sperl, and
  Lampert}{Rebuffi et~al.}{2017}]{rebuffi2017icarl}
Rebuffi, Sylvestre-Alvise, Alexander Kolesnikov, Georg Sperl, and Christoph~H
  Lampert (2017).
\newblock icarl: Incremental classifier and tation learning.
\newblock In {\em Conference on Computer Vision and Pattern Recognition}.

\bibitem[\protect\citeauthoryear{Riemer, Cases, Ajemian, Liu, Rish, Tu, and
  Tesauro}{Riemer et~al.}{2019}]{riemer2018learning}
Riemer, Matthew, Ignacio Cases, Robert Ajemian, Miao Liu, Irina Rish, Yuhai Tu,
  and Gerald Tesauro (2019).
\newblock Learning to learn without forgetting by maximizing transfer and
  minimizing interference.
\newblock {\em International Conference on Learning Representations\/}.

\bibitem[\protect\citeauthoryear{Schmidhuber}{Schmidhuber}{1987}]{schmidhuber1987evolutionary}
Schmidhuber, Jurgen (1987).
\newblock {\em Evolutionary principles in self-referential learning, or on
  learning how to learn}.
\newblock Ph.\ D. thesis, Institut fur Informatik,Technische Universitat
  Munchen.

\bibitem[\protect\citeauthoryear{Shin, Lee, Kim, and Kim}{Shin
  et~al.}{2017}]{shin2017continual}
Shin, Hanul, Jung~Kwon Lee, Jaehong Kim, and Jiwon Kim (2017).
\newblock Continual learning with deep generative replay.
\newblock In {\em Advances in Neural Information Processing Systems}.

\bibitem[\protect\citeauthoryear{Sutton}{Sutton}{1990}]{sutton1990integrated}
Sutton, Richard (1990).
\newblock {Integrated architectures for learning planning and reacting based on
  approximating dynamic programming}.
\newblock In {\em International Conference on Machine Learning}.

\bibitem[\protect\citeauthoryear{Zenke, Poole, and Ganguli}{Zenke
  et~al.}{2017}]{zenke2017continual}
Zenke, Friedemann, Ben Poole, and Surya Ganguli (2017).
\newblock Continual learning through synaptic intelligence.
\newblock In {\em International Conference on Machine Learning}.

\end{thebibliography}
	\clearpage
	
	\appendix
	\section*{Appendix}

	\section{Discussion on the Connection to Few-Shot Meta-Learning} 
	\label{fixed_rep}
Our approach is different from gradient-based meta-learning in two ways; first, we only update PLN during the inner updates whereas maml (and other gradient-based meta-learning techniques) update all the parameters in the inner update. By not updating the initial layers in the inner update, we change the optimization problem from "finding a model initialization with \textit{xyz} properties" to "finding a model initialization and learning a fixed representation such that starting from the learned representation it has \textit{xyz} properties." This gives our model freedom to transform the input into a more desirable representation for the task---such as a sparse representation.

\begin{figure}
	\centering
	\includegraphics[width=0.99\linewidth]{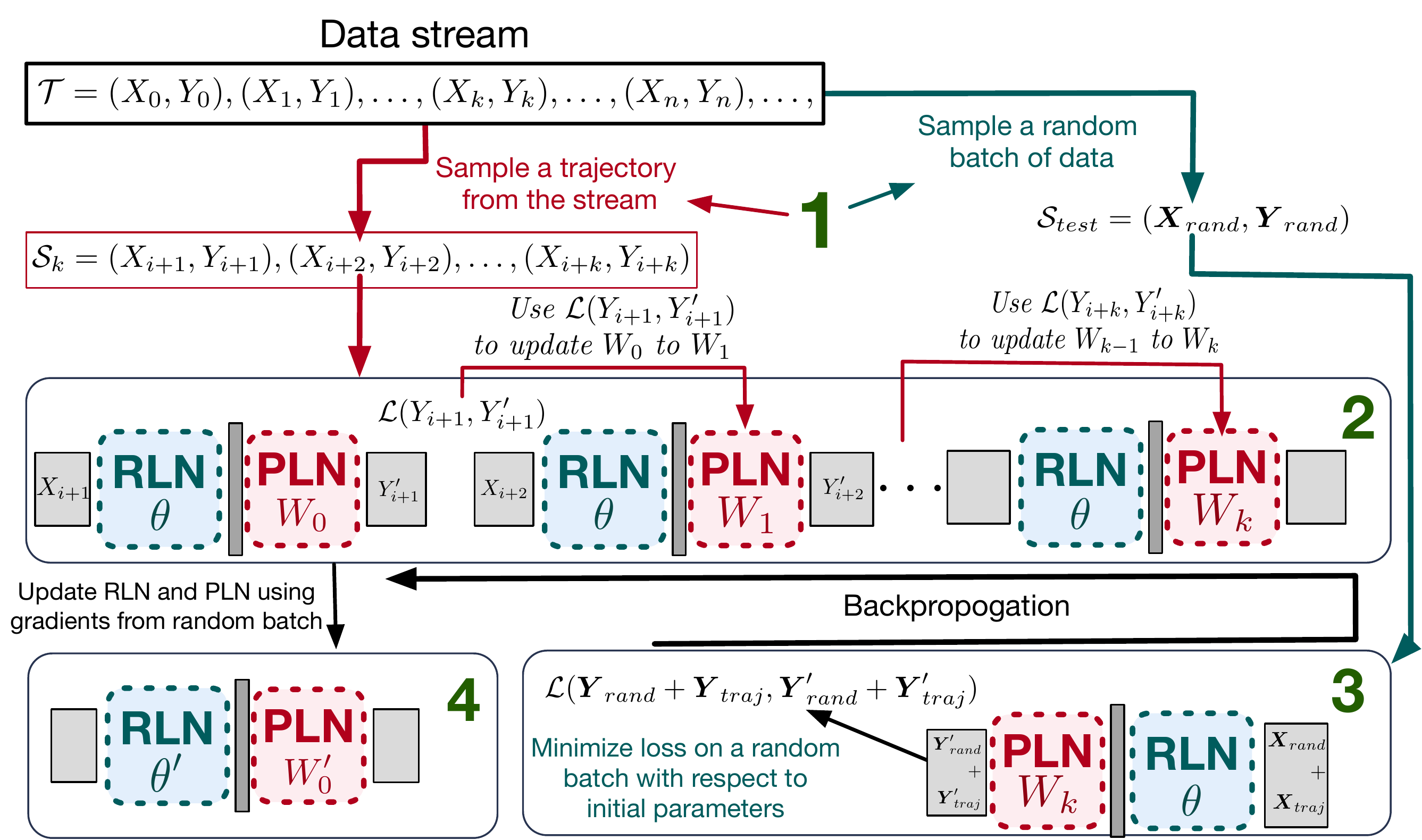}
	\caption{Flowchart elucidating a single gradient update for representation learning. (1) We sample trajectory $\mathcal S_k$ from our stream of data for inner updates in the meta-training, and another trajectory $\mathcal S_{test}$ for evaluation. (2) We use  $\mathcal S_k$ to do $k$ gradient updates on the PLN (Prediction learning network). (3) We then use this updated network to compute loss on the $S_k + S_{test} $ and compute gradients for this loss with respect to the initial parameters $\theta_1, W_1$. (4) Finally, we update our initial parameters $\theta, W_0$ to $\theta', W'_0$.}		
	\label{fig:update}
\end{figure}
\begin{table}
	\caption{Parameters for Sinusoidal Regression Experiment}
	\label{hyper_sin}
	\centering
	\begin{tabular}{lll}
		\toprule
		Parameter     & Description     & Value \\
		\midrule
		Meta LR & Learning rate used for the meta-update  & 1e-4     \\
		Meta Update Optimizer     & Optimizer used for the meta-update & Adam      \\
		Inner LR     & LR used for the inner updates for meta-learning & 0.003     \\
		Inner LR Search    & Inner LRs tried before picking the best & [0.1, 1e-6]  \\
		Steps-per-function     & Number of gradient updates for each of the ten tasks      & 40  \\
			Inner steps    & Number of inner gradient steps    & 400  \\
		Total layers     & Total layers in the fully connected NN     &  9  \\
		Layer Width     & Number of neurons in each layer    &  300  \\
		Non-linearly     & Non-linearly used   & relu \\
		RLN Layers     & Number of layers used for learning representation    &  6  \\
		Pre-training set    & Number of functions in the meta-training set    &  400  \\
		\bottomrule
	\end{tabular}
\end{table}

Secondly, we sample trajectories and do correlated updates in the inner updates, and compute the meta-loss with respect to a batch of data representing the CLP problem at large. This changes the optimization from "finding an initialization that allows for quick adaptation" (such as in maml \cite{Finn:EECS-2018-105}) to "finding an initialization that minimizes interference and maximizes transfer." Note that we learn the RLN and the initialization for PLN using a single objective in an end-to-end manner. 

We empirically found that having an RLN is extremely important for effective continual learning, and vanilla maml trained with correlated trajectories performed poorly for online learning.
	\section{Reproducing Results}
	\label{app_exp}

We release our code, and pretrained \OML\ models for Split-Omniglot and Incremental Sine Waves available at \url{https://github.com/Khurramjaved96/mrcl}. In addition, we also provide details of hyper-parameters used from learning the representations of Incremental Sine Waves experiment and Split-Omniglot in Table \ref{hyper_sin} and \ref{hyper_omni} respectively.  

For online learning experiments in Figure \ref{fig:regression} and \ref{fig:classification}, we did a sweep over the only hyper-parameter, learning rate,  in the list [0.3, 0.1, 0.03, 0.01, 0.003, 0.001, 0.0003, 0.0001, 0.00003, 0.00001] for each method on a five validation trajectories and reported result for the best learning rate on 50 random trajectories.

	\begin{table}
	\caption{Parameters for Omniglot Representation Learning}
	\label{hyper_omni}
	\centering
	\begin{tabular}{lll}
		\toprule
		Parameter     & Description     & Value \\
		\midrule
		Meta LR & Learning rate used for the meta-update  & 1e-4     \\
		Meta update optimizer     & Optimizer used for the meta-update & Adam      \\
		Inner LR     & LR used for the inner updates for meta-learning & 0.03     \\
		Inner LR Search    & Inner LRs tried before picking the best & [0.1, 1e-6]  \\
			Inner steps    & Number of inner gradient steps    & 20  \\
		Conv-layers     & Total convolutional layers    &  6  \\
			FC Layers    & Total fully connected layers    &  2  \\
			RLN    & Layers in RLN    &  6  \\
		Kernel     & Size of the convolutional kernel   &  3x3 \\
		Non-linearly     & Non-linearly used   & relu \\
		Stride    & Stride for convolution operation in each layer    &  [2,1,2,1,2,2]  \\
			\# kernels		     & Number of convolution kernels in each layer    &  256 each \\
	Input    & Dimension of the input image   &  84 x 84  \\
		\bottomrule
	\end{tabular}
\end{table}
	\begin{figure}
		\centering
		\includegraphics[width=0.80 \linewidth]{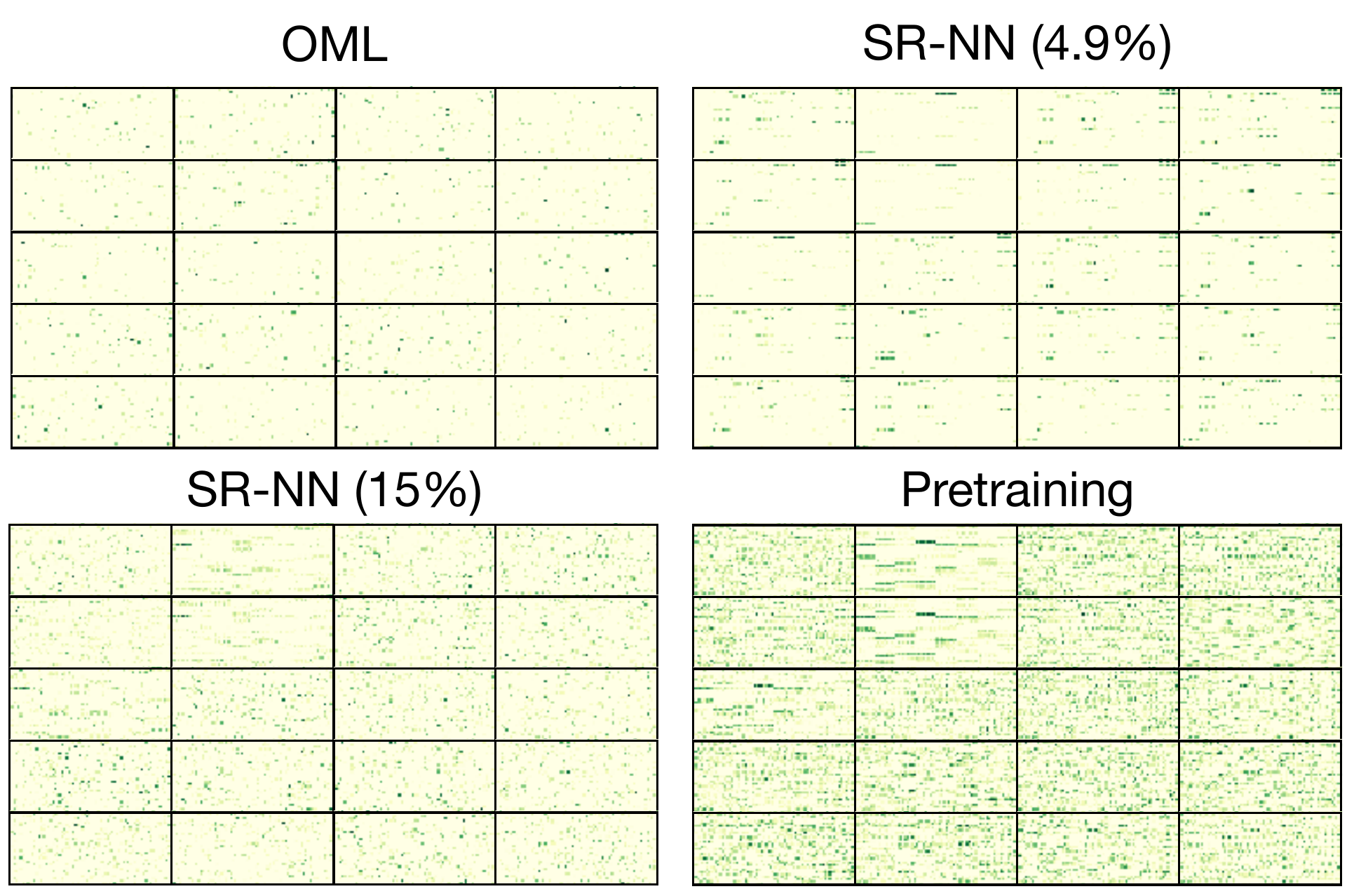}
		\caption{More samples of representations for random input images for different methods. Here SR-NN (4.9\%) is trained to have similar sparsity as \OML\ whereas SR-NN (15\%) is trained to have the best performance on Split-Omniglot benchmark.}
		\label{fig:combined_rep}
	\end{figure}
	\subsection{Computing Infrastructure}
	
\begin{figure}

	\begin{center}
		\includegraphics[width=0.50\textwidth]{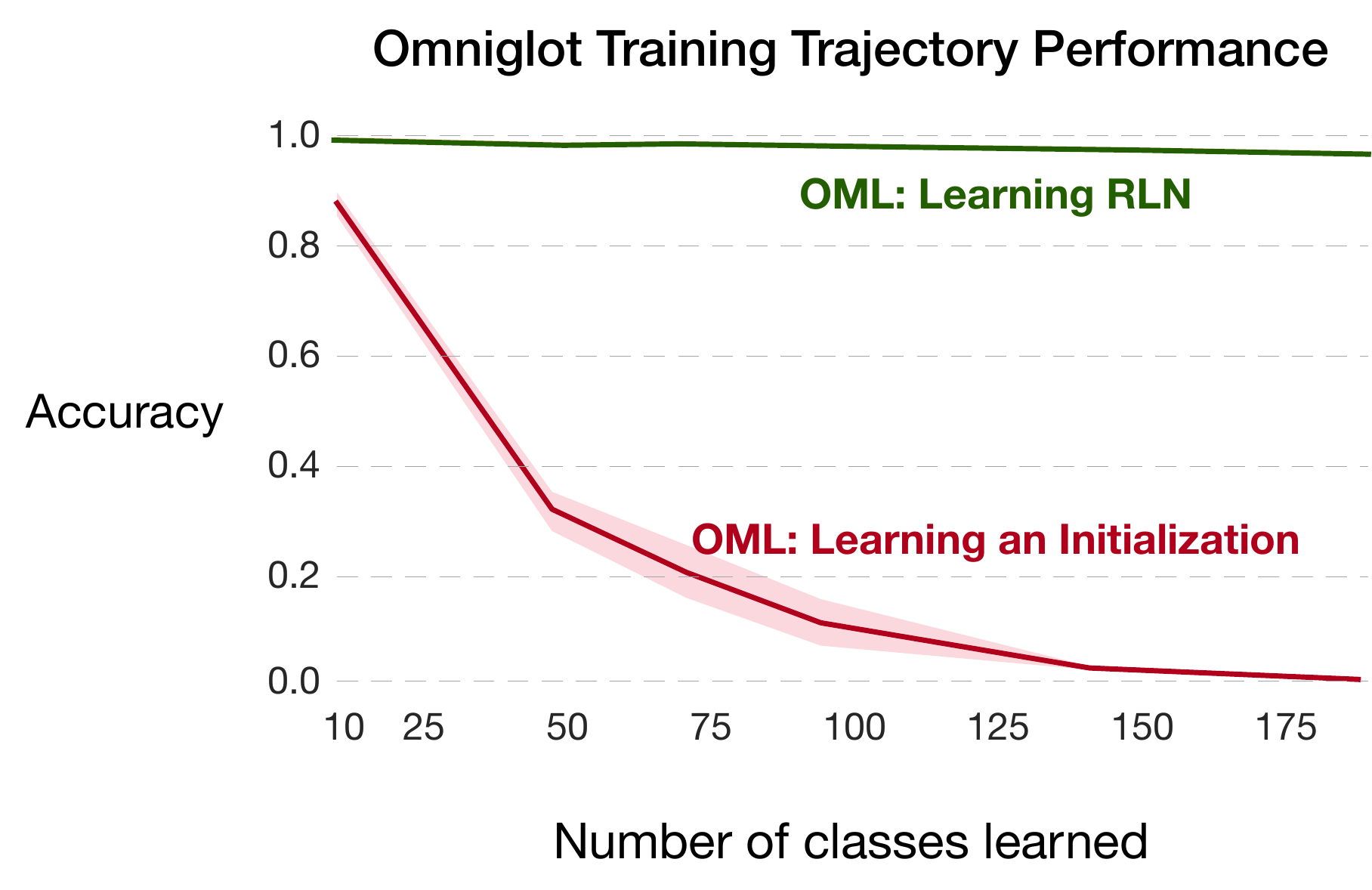}
	\end{center}
	\caption{Instead of learning an encoder $\encoder$ , we learn an initialization by updating both $\theta$ and $W$ in the inner loop of meta-training. In "\OML\ without RLN," we also update both at meta-test time whereas in "\OML\ without RLN at test time," we fix $theta$ at meta-test time just like we do for \OML\ . For each of the methods, we report the training error during meta-testing. It's clear from the results that a model initialization is not an effective bias for incremental learning. Interestingly, "\OML\ with RLN at test time" doesn't do very poorly. However, if we know we'll be fixing $\theta$ at meta-test time, it doesn't make sense to update it in the inner loop of meta-training (Since we'd want the inner loop setting to be as similar to meta-test setting as possible.}
	\label{RLN}
\end{figure}

We learn all representations on a single V100 GPU; even with a deep neural network and meta-updates involving roll-outs of length up to 400, \OML\ can learn representations in less than five hours for both the regression problem and omniglot experiments. For smaller roll-outs in Omniglot, it is possible to learn good representations with-in an hour. Note that this is despite the fact that we did not use batch-normalization layers or skip connections which are known to stabilize and accelerate training.
	\section{Representations}
	\label{rep_appendix}
\begin{figure}
	\centering
	\includegraphics[width=0.40\linewidth]{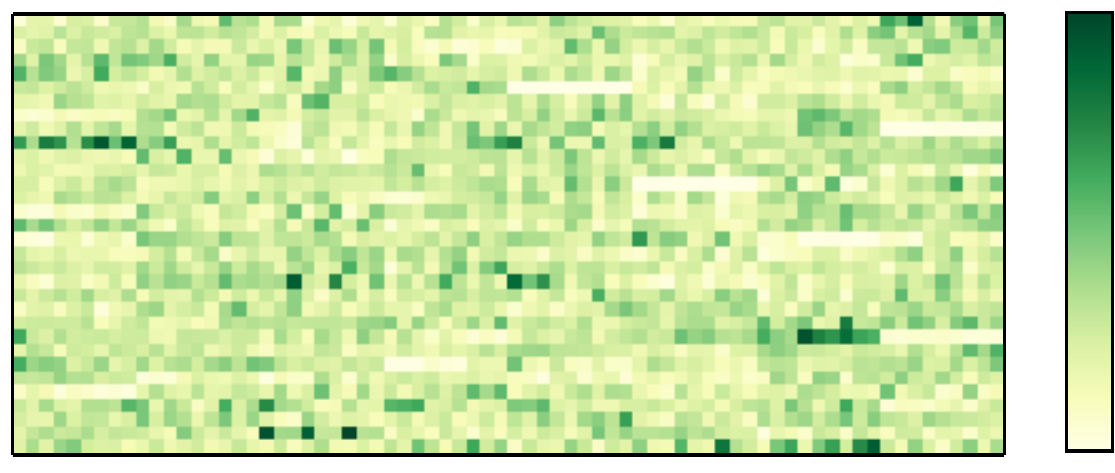}
	\caption{Average activation map for the best performing SR-NN with 15\% sparsity. Scale goes  from 0 to max (Light to dark green.)}
	\label{fig:good_srnn}
\end{figure}

We present more samples of the learned representations in Figure \ref{fig:combined_rep}. We also include the averaged representation for the best performing SR-NN model (15\% instance sparsity) in Figure \ref{fig:good_srnn} which was excluded from Figure \ref{fig:rep_comparison} due to lack of space. 

% \section{Learning Initializations for PLN}
% In this paper, we have focused on learning representations using meta-learning. However our objective does more than that: in addition to learning a Representation Learning Network (RLN), it also learns an initialization for Prediction Learning Network (PLN) which is suitable for learning from the learned representation. This initialization is similar to what maml \citep{finn2017model} does for few-shot-learning. All the results in the paper use this initialization for continual learning, however, it is not clear if the initialization of PLN plays an important role in continual learning. 

% To empirically answer this question, we compare \OML\ to $\OML_{random\_init}$, which is same as \OML\ except we initialize PLN with random weights. As it can be seen in Figure \ref{fig:init_effect}, the learned initialization of PLN results in a measurable difference in performance. In-fact, with random weights, the overall error nearly doubles for the incremental sine function problem. However, it might be possible to    
% \begin{figure}
% 	\centering
% 	\includegraphics[width=0.99\linewidth]{figures/effect_of_init}
% 	\caption{Our objective learns representations using RLN as well as network initialization for PLN for effective continual learning. The performance of \OML\ is noticeably worse if we initialize PLN using random weights instead of the weights learned by our OML objective.}
% 	\label{fig:init_effect}
% \end{figure}

\begin{algorithm}[H]
	\centering
	\caption{Meta-Training : Approximate Implementation of the $\OML$ Objective}\label{algorithm_approx}
	\begin{algorithmic}[1]
		\REQUIRE $p(\task)$: distribution over tasks
		\REQUIRE $\alpha$, $\beta$: step size hyperparameters
		\REQUIRE $m$: No of inner gradient steps per update before truncation
		\STATE randomly initialize $\repparams, \taskparams$
		\WHILE{not done}
		\STATE Sample task $\task_i \sim p(\task)$
		\STATE Sample $\mathcal S_{train}^i$ from $p(\mathcal S_k| \task_i)$
		\STATE $\taskparams_0 = \taskparams$
		\STATE $\nabla_{accum} = \mathbf{0}$
		\WHILE {$j \le |\mathcal S_{train}$|}
		\FOR{$k$ in $1, 2, \dots , m$}
		\STATE  $\taskparams_j=\taskparams_{j-1}-\alpha
		\nabla_{\taskparams_{j-1}}  \loss_i(f_{\repparams, \taskparams_{j-1}}(X_{j}^i), Y_j^i)$ 
		\STATE $j$ = $j + 1$
		\ENDFOR
		\STATE Sample $S_{test}^i$ from $p(\mathcal S_k| \task_i)$
		\STATE $\repparams = \repparams +   \nabla_\repparams \loss_i (f_{\repparams, \taskparams_{j}}[S_{test}[0:j, 0]], S_{test}^i[0:j,1])$
		\STATE Stop Gradients$(f_{\repparams, \taskparams_{j}}))$
		\ENDWHILE
		\ENDWHILE
	\end{algorithmic}
\end{algorithm}

\subsection{Why Learn an Encoder Instead of an Initialization}
\label{initvsrep}
We empirically found that learning an encoder results in significantly better performance than learning just an initialization as shown in Fig \ref{RLN}. Moreover, the meta-learning optimization problem is more well-behaved when learning an encoder (Less sensitive to hyper-parameters and converges faster). One explanation for this difference is that a global and greedy update algorithm -- such as gradient descent -- will greedily change the weights of the initial layers of the neural network with respect to current samples when learning on a highly correlated stream of data. Such changes in the initial layers will interfere with the past knowledge of the model. As a consequence, an initialization is not an effective inductive bias for incremental learning. When learning an encoder $\encoder$, on the other hand, it is possible for the neural network to learn highly sparse representations which make the update less global (Since weights connecting to features that are zero remain unchanged).  

\end{document}